\documentclass[conference]{IEEEtran}
\IEEEoverridecommandlockouts
\usepackage{comment}
\usepackage{cite}
\usepackage{amsmath,amssymb,amsfonts}
\usepackage{algorithmic}
\usepackage{algorithm}
\usepackage{graphicx}
\usepackage{textcomp}
\usepackage{xcolor}
\usepackage{booktabs}
\usepackage{url}
\usepackage{hyperref}

\def\BibTeX{{\rm B\kern-.05em{\sc i\kern-.025em b}\kern-.08em
    T\kern-.1667em\lower.7ex\hbox{E}\kern-.125emX}}

\begin{document}

\title{Self-Paced Curriculum Reinforcement Learning\\
for Autonomous Superbike Racing in Simulation}

\author{
  \IEEEauthorblockN{Luca Ghisi}
  \IEEEauthorblockA{
    \textit{Università degli Studi di Milano}\\
    Milan, Italy}
  \and
  \IEEEauthorblockN{Jacopo Essenziale}
  \IEEEauthorblockA{\textit{Funny Tales}\\
    Milan, Italy}
  \and
    \IEEEauthorblockN{Carlo D'Eramo}
  \IEEEauthorblockA{
    \textit{ University of Würzburg}\\
    Würzburg, Germany}
  \and
 \IEEEauthorblockN{Matteo Luperto}
  \IEEEauthorblockA{
    \textit{Università degli Studi di Milano}\\
    Milan, Italy\\ matteo.luperto@unimi.it}
}

\maketitle

\begin{abstract}
Autonomous Racing has seen remarkable progress through deep Reinforcement Learning (RL), primarily for four-wheeled vehicles. However, motorbikes introduce substantially greater complexity due to the need to manage balance and lean angle, in addition to more reactive steering and throttle control, and a smaller weight. In this work, we present a framework for training an autonomous agent to race a superbike in VRider SBK, a physics-accurate Unity-based motorbike simulator. Our approach integrates Soft Actor-Critic (SAC) with Self-Paced curriculum Deep reinforcement Learning (SPDL), which dynamically generates progressively more challenging tasks based on the agent's performance — without requiring manual curriculum design. The agent's state space comprises proprioceptive features extended with lean-angle history, along with global track features via course points. The reward signal is shaped to encourage progress along the track while penalizing instability-inducing behaviors specific to two-wheeled dynamics. Preliminary experimental results demonstrate that SPDL outperforms SAC alone in training efficiency, lap time, and driving stability across multiple tracks and motorbike models, establishing a first baseline for RL-based autonomous motorbike racing.
\end{abstract}


\section{Introduction}

Autonomous Racing is a demanding subfield of autonomous driving, requiring an agent to perceive its environment, plan an optimal trajectory, and execute precise vehicle control at the physical limits of the vehicle. In recent years, end-to-end deep Reinforcement Learning (RL) has emerged as a compelling alternative to classical modular pipelines — composed of perception, planning, and control modules — showing super-human performance in high-fidelity simulators such as Gran Turismo Sport and Gran Turismo 7 \cite{fuchs2021super,wurman2022outracing,vasco2024super1,lee2025}.

\begin{figure}[!t]
\begin{center}
    \includegraphics[width=\linewidth]{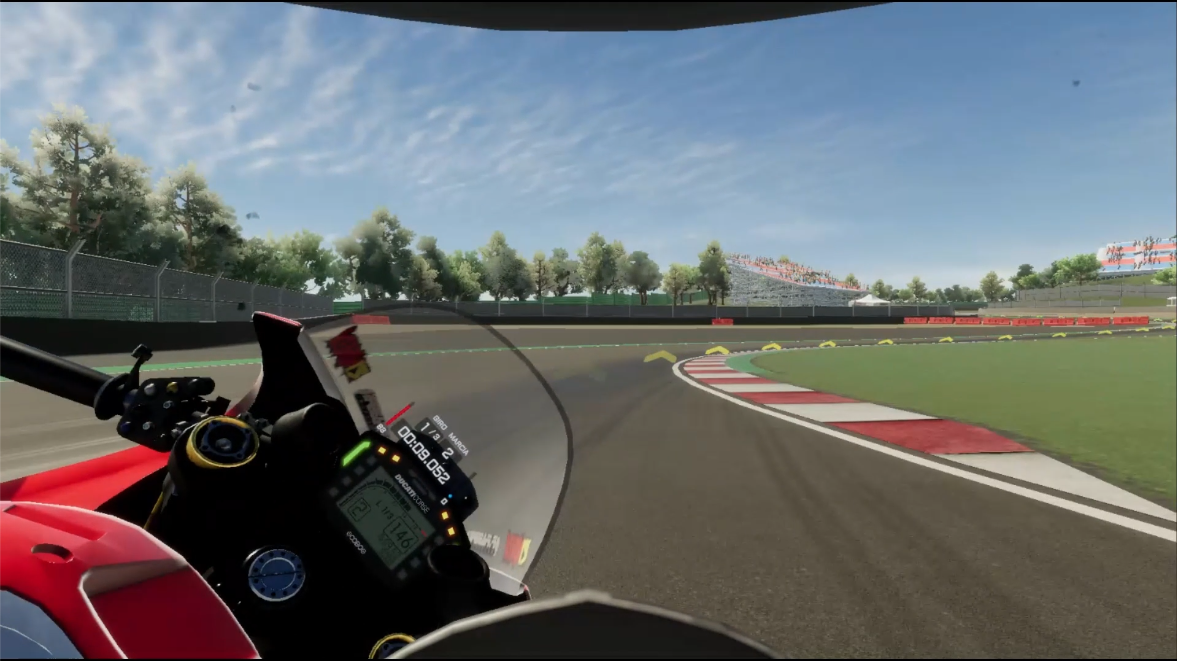}
    \includegraphics[width=\linewidth]{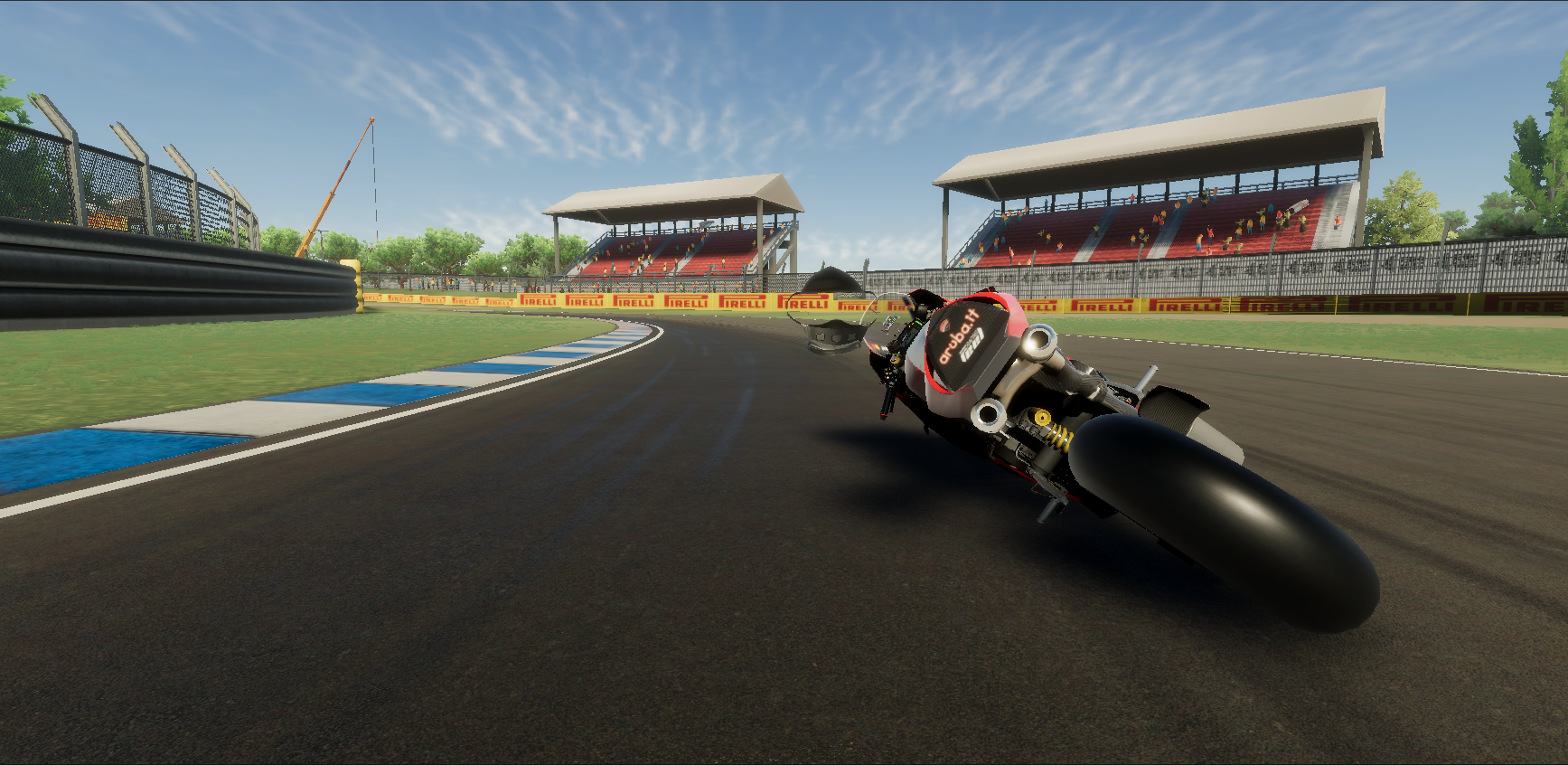}
    \caption{
    An example of the view from the VRider SBK game (top) and a rendering of the game from a third-person point of view (bottom), showing the motorbike leaning through a curve at high speed.
    }
    \label{fig:moto}
\end{center}
\end{figure}

Despite significant progress for four-wheeled vehicles, motorbike racing remains largely unexplored in the RL literature. A racing motorbike introduces fundamentally different physical and dynamical challenges compared to a race car: besides steering and throttle/brake control, the agent must actively manage the \emph{lean angle}, which is essential to counteract centrifugal forces when cornering, as can be seen in Fig. \ref{fig:moto}. Failure to control the lean angle leads to loss of balance and falling — a failure mode absent in four-wheeled vehicles. At the same time, racing bikes are subject to stronger dynamics in terms of breaking and acceleration than most racing car equivalents, due to a higher power to weight ratio.

In this paper, we present a framework for training an autonomous RL agent to race a superbike in VRider SBK\footnote{\url{https://vridergame.com/}}, a realistic motorbike simulator modelling real-world Superbike World Championship tracks and physics. Our contributions are:
\begin{itemize}
  \item A Reinforcement Learning architecture tailored to two-wheeled vehicles, including a rich state space with lean-angle history and a reward signal designed to handle motorbike-specific instabilities.
  \item Integration of Self-Paced curriculum Deep reinforcement Learning (SPDL) \cite{klink2020self,klink2021probabilistic,klink2022curriculum} with Unity ML-Agents, enabling dynamic and automatic curriculum generation within a game simulation.
  \item Preliminary experimental evaluation demonstrating that SPDL achieves faster convergence, lower lap times, and more stable driving compared to SAC alone, with results generalized across multiple tracks and motorbike models.
\end{itemize}

\begin{figure}[!t]
\begin{center}
    \includegraphics[width=\linewidth]{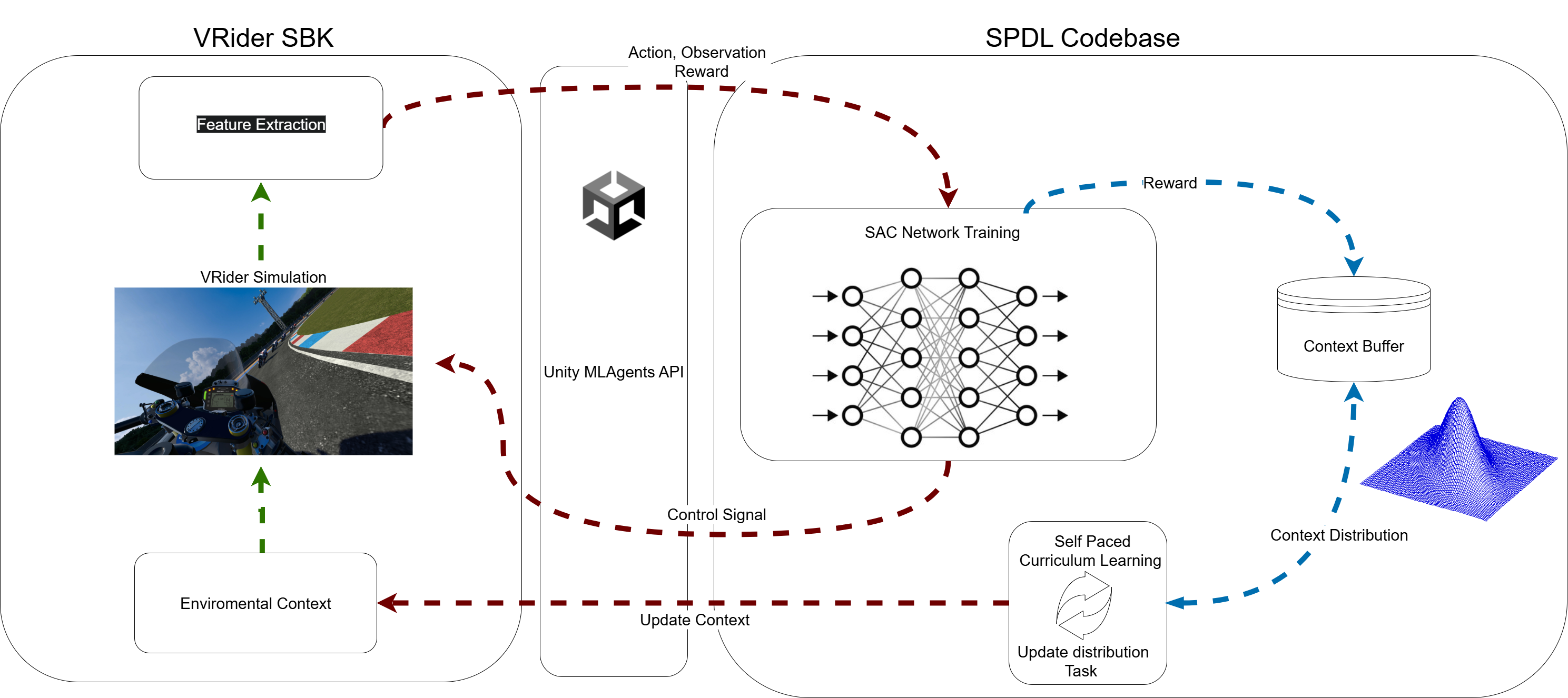}
    \caption{
    The architecture of our system.
    }
    \label{fig:Architecture}
\end{center}
\end{figure}

\section{Related Work}

\subsection{Reinforcement Learning for Autonomous Racing}

The application of RL to autonomous racing has advanced rapidly. Fuchs~\emph{et al.}~\cite{fuchs2021super} achieved super-human lap times in Gran Turismo Sport using SAC, introducing a dense progress-based reward and kinetic-energy wall-contact penalty. Song~\emph{et al.}~\cite{song2021autonomous2} extended this to overtaking via a manually designed three-stage curriculum. Wurman~\emph{et al.}~\cite{wurman2022outracing} introduced GT Sophy, leveraging QR-SAC with complementary training across diverse environment configurations to produce a champion-level racing agent. More recently, Vasco~\emph{et al.}~\cite{vasco2024super1} and Lee~\emph{et al.}~\cite{lee2025} incorporated vision-based inputs with asymmetric actor-critic architectures. Generalization across vehicles and conditions was addressed by Grooten~\emph{et al.}~\cite{grooten2025out} via the SPARC algorithm. All existing work, however, focuses exclusively on four-wheeled vehicles.

\subsection{Self-Paced Deep Reinforcement Curriculum Learning}

Self-Paced Deep reinforcement Learning (SPDL) \cite{klink2020self,klink2021probabilistic,klink2022curriculum} extends contextual RL by automatically generating task distributions of increasing difficulty based on the agent's current performance, rather than relying on manually designed task sequences \cite{song2021autonomous2}. The algorithm maintains a context distribution $p(c|\nu)$ over a contextual MDP and optimizes:
\begin{equation}
  \max_{\nu} \; \mathbb{E}_{p(c|\nu)}\bigl[J(\pi, c)\bigr] - \alpha \, D_{\mathrm{KL}}\!\bigl(p(c|\nu) \| \mu(c)\bigr)
  \label{eq:SPDL}
\end{equation}
where $\mu(c)$ is the target task distribution and $\alpha$ controls the trade-off between reward maximization and convergence to the target. A KL divergence constraint prevents abrupt jumps between consecutive distributions, ensuring stable curriculum progression.

\section{Problem Formulation and Simulation}

VRider SBK is a Unity-based superbike simulator developed by Funny Tales, featuring 14 real-world Superbike championship tracks and five bike models (e.g., Ducati Panigale V4~R, Kawasaki Ninja ZX-10RR). The game is designed for VR, thus its physics is accurately modelled to reduce uncanny experience and sickness by users. The simulator provides rich data through providers for vehicle state (position, velocity, rotation), track data (median and ideal splines, track limits, checkpoints), collision and fall events, and lap timing. The integration with RL is realised through Unity ML-Agents \cite{juliani2020}, combined with a Gym wrapper \cite{gym} and Stable Baselines 3 \cite{raffin2021stable}.  An overview of the system is provided in Fig. \ref{fig:Architecture}.


\subsection{State Space}

The agent's observation vector combines proprioceptive and global features, all normalized to $[-1, 1]$.

\textbf{Proprioceptive features:} linear velocity (with longitudinal component), angular velocity, current and previous steering and lean angles, lean-angle $\theta$, tyre slip ratios, previous control actions, and orientation relative to the followed trajectory.

\textbf{Global features:} 60 equally-spaced 3D course points sampled along the current trajectory (center line and left/right track limits), computed via Catmull-Rom spline interpolation at the agent's current speed. This gives the agent a dynamic, speed-adapted look-ahead of the track shape.

The inclusion of lean-angle history and its deltas is motivated by the dynamics of motorbike control: the agent must anticipate and suppress oscillatory lean behavior that can lead to loss of control.

\subsection{Action Space}

Following the autonomous racing literature \cite{fuchs2021super}, two continuous actions in $[-1, 1]$ are used: (i) steering, which also drives the lean angle of the bike, and (ii) a combined throttle/brake signal, encoding the constraint that both are not applied simultaneously.

\section{Reward Design}

The reward signal must be dense and guide the agent to minimize lap time while managing the instability-prone dynamics of a two-wheeled vehicle. The total reward at each timestep is:
\begin{equation}
\begin{aligned}
  r &=  \lambda_{\mathrm{way}} \lambda_{prog}* r_{prog} + \lambda_o * r_{off} + \lambda_w* r_{wall} + \lambda_s * r_{steer} + \\ 
    &\lambda_b* r_{lean}
		\quad + \lambda_{av} * r_{AngVel} + \lambda_{h_s}* r_{h_s} + \\ &\lambda_{h_l}* r_{h_l} + \lambda_{maxl}* r_{maxl} 
		\quad + \lambda_{back} * r_{backwards} + \\ 
        &\lambda_{still} *r_{still}  + \lambda_{slip} *r_{slip}
        \end{aligned}
  \label{eq:reward}
\end{equation}
\noindent
where $\lambda_*$ are hyperparameters that control the influence of each reward term and act as weights.

\textbf{Course progress} $r_{\mathrm{prog}} = \mathrm{cp}_t - \mathrm{cp}_{t-1}$: the displacement along the followed trajectory between consecutive timesteps ($\Delta t = 0.2$~s), following \cite{fuchs2021super}. A wrong-way multiplier $\lambda_{\mathrm{way}}=-1$ negates this reward when the bike's heading deviates more than 90° from the track tangent (and is equal to $1$ otherwise).

\textbf{Off-track penalty} $r_{\mathrm{off}} = -(o_t - o_{t-1})|v_t|$: penalizes time spent with a tyre outside the track limits, weighted by speed.

\textbf{Wall penalty} $r_{\mathrm{wall}} = -(w_t - w_{t-1})|v_t|$: kinetic-energy-based wall contact penalty \cite{fuchs2021super}.

\textbf{Stability penalties}: Steering ($r_{\mathrm{steer}}$), leaning ($r_{\mathrm{lean}}$), and angular velocity $AngVel$ penalties each penalize large angle changes between consecutive steps ($-|\theta_t - \theta_{t-1}|$). A history $h$ penalty based on a sigmoid function of accumulated angle deltas discourages oscillatory behavior over recent steps and is used for steering ($h_s$) and leaning ($h_l$). A max-leaning $maxl$ penalty prevents extreme lean exploitation. Tyre $slip$, $standstill$, and $backward$ motion penalties complete the set. Episode failure (reset) is triggered if any penalty timer expires or if the bike falls.

\section{Self-Paced Curriculum and Context Variables}
\begin{figure}[!t]
\begin{center}
    \includegraphics[angle=90, width=0.85\linewidth]{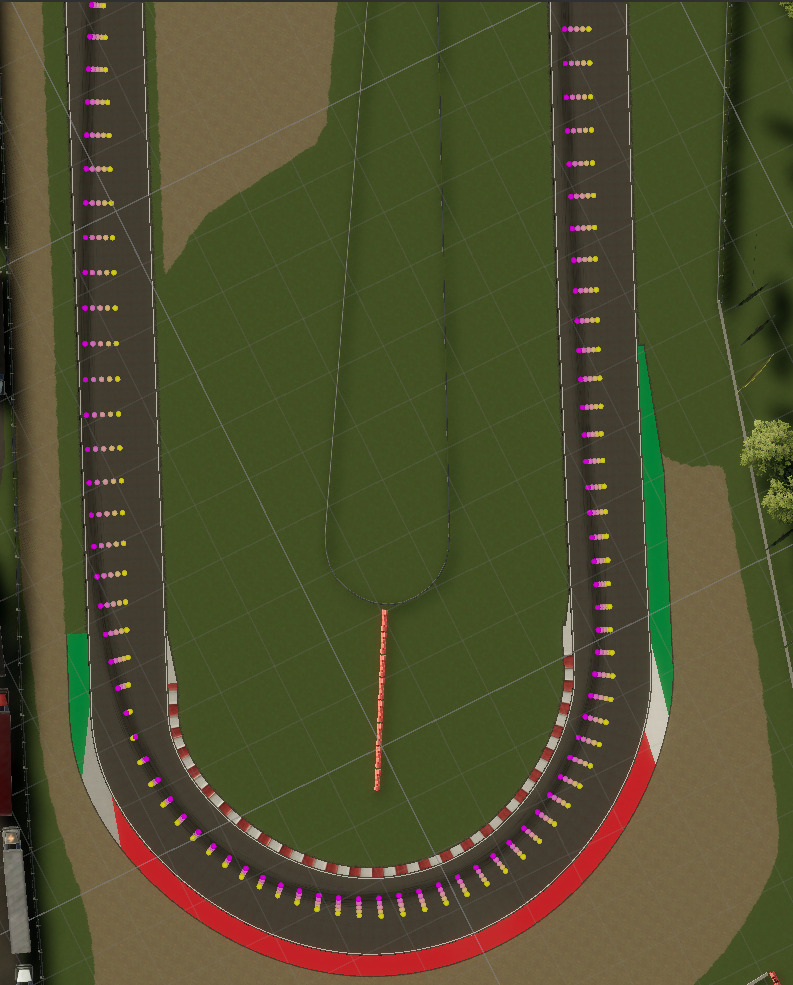}
    \includegraphics[width=0.85\linewidth]{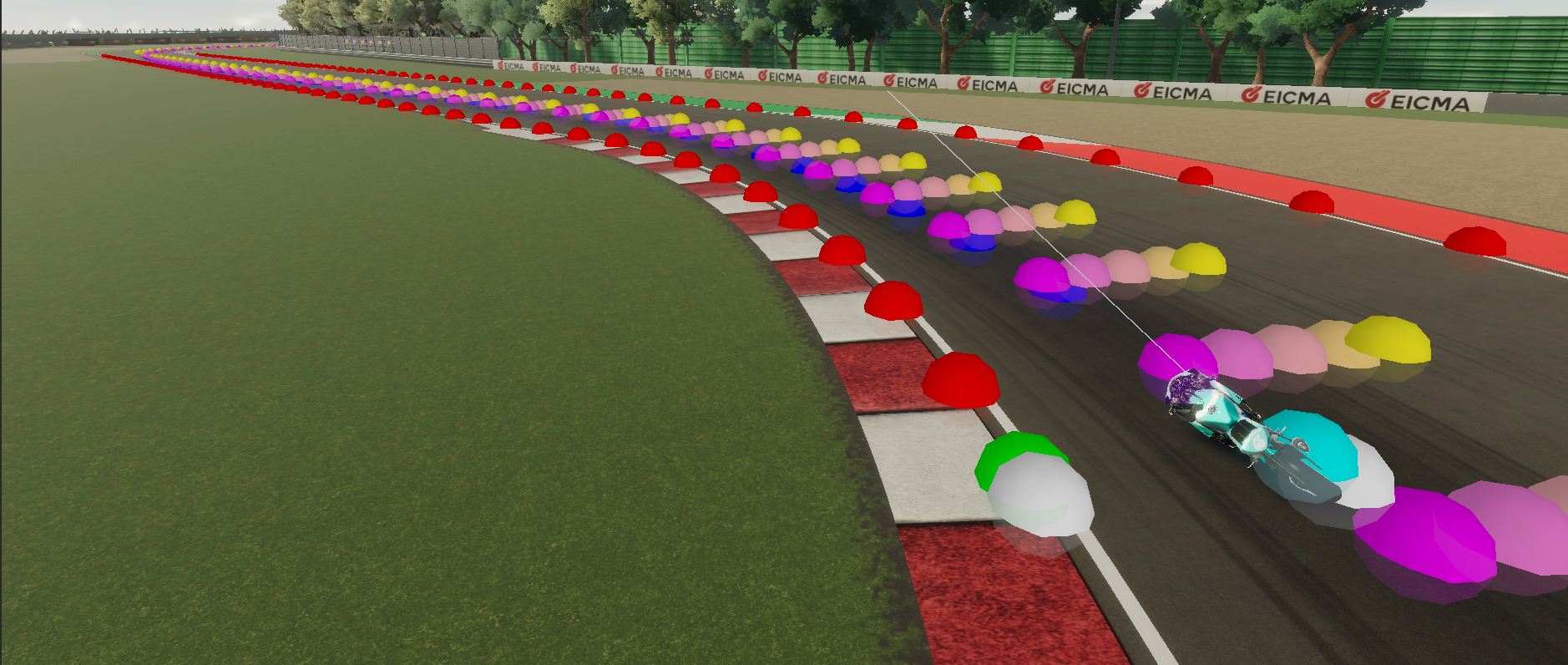}
    \caption{
    (top) The trajectories followed by the motorbike, from the \emph{median} track (yellow) to the ideal one (purple). (bottom) A close-up of the motorbike following the trajectories.
    }
    \label{fig:context}
\end{center}
\end{figure}

The key insight of our curriculum is that the trajectory the agent follows during training can serve as the difficulty axis: starting from the track \emph{median} (centerline) and progressing toward the \emph{ideal} racing line (optimal in-out-in trajectory for minimum lap time). 

Three context variables define the current task:
\begin{itemize}
  \item $c_1$: Distance from the median trajectory (higher = harder).
  \item $c_2$: Distance from the ideal trajectory (lower = harder).
  \item $c_3$: Maximum speed limit.
\end{itemize}

The context triple $(c_1, c_2, c_3)$ determines which interpolated spline the agent should follow via:
\begin{equation}
  \alpha = \frac{1}{2}\left(\frac{c_1}{c_1 + c_2} + \frac{c_3}{v_{\max}}\right)
   \label{eq:alpha}
\end{equation}
where $\alpha \in [0,1]$ is the interpolation parameter used in Eq \ref{eq:SPDL}.  The first term $\frac{c_1}{c_1 + c_2}$ combines the distances to indicate which spline is closest, while $\frac{c_3}{v_{\max}}$ normalises the speed; both are weighted equally for spline selection.
Initially, all variables are set near the median trajectory (easy task). As the agent's performance exceeds a threshold $\delta$, SPDL automatically shifts the distribution toward the target context (ideal line at full speed). The KL constraint $D_{\mathrm{KL}}(p(c|\nu) \| p(c|\nu')) \leq \epsilon$ prevents abrupt transitions. We use SPDL version 2 \cite{klink2021probabilistic}, which requires setting only a single performance threshold $\delta$, making it significantly easier to tune than the two-parameter version \cite{klink2020self}.

\section{Experimental Evaluation}

\subsection{Setup}

All experiments use SAC from Stable Baselines 3 (learning rate $2 \times 10^{-5}$, replay buffer $3 \times 10^6$, batch size 512), with 10 parallel environments. Two algorithms are compared: \textbf{SAC} (baseline, trained on the ideal trajectory throughout) and \textbf{SPDL} (SAC + Self-Paced Curriculum). Training runs for up to 1900 iterations of 5000 steps each ($\approx$9.5M environment steps per run). The primary track is \textbf{Barcelona}; additional evaluation uses \textbf{Cremona} and \textbf{Phillip Island}.  All experiments use the Ducati Panigale V4~R unless stated otherwise, in Hot Lap (time-trial) mode. Parameters $\lambda$ are set by expert knowledge.



\subsection{Deployment Results}

\begin{table}[t]
\centering
\caption{Average performance metrics on Barcelona (4 sessions per model). \small ``—'' = No lap completed (bike unable to complete the track).}
\label{tab:barcelona}
\begin{tabular}{@{}lcccc@{}}
\toprule
Metric & \multicolumn{2}{c}{Iter.~1150} & \multicolumn{2}{c}{Iter.~1900} \\
\cmidrule(lr){2-3}\cmidrule(lr){4-5}
 & SPDL & SAC & SPDL & SAC \\
\midrule
Avg. lap time       & 1:32.85 & —       & \textbf{1:30.75} & 1:31.32 \\
Dist. to ideal (m)  & 2.99    & 4.59    & 3.15             & 2.89    \\
Off-track time (s)  & 1.74    & 9.22    & \textbf{1.17}    & 2.15    \\
Falls               & 0       & 3       & 0                & 0       \\
Collisions          & 0       & 0       & 0                & 0       \\
\bottomrule
\end{tabular}

\end{table}

Table~\ref{tab:barcelona} shows the key deployment metrics for the Barcelona circuit. At iteration 1150, the SAC-only model fails to complete a single lap, repeatedly losing control between the first and second track sectors. SPDL completes all laps with zero falls and an average time of 1:32.85. By iteration 1900, SAC has trained sufficiently to complete laps, with a best of 1:31.32; SPDL achieves 1:30.75 — a gain of 0.57~s. Over three consecutive laps, SPDL gains 0.77~s, accumulating advantage primarily in the initial part of the circuit.

\begin{table}[t]
\centering
\scriptsize
\caption{Track generalization with SPDL after 1150 iterations.}
\label{tab:generalization}
\begin{tabular}{@{}lcccc@{}}
\toprule
Track & Avg. Lap & Dist. to ideal (m) & Off-track (s) & Avg. Lap (SAC)\\
\midrule
Barcelona    & 1:32.85 & 2.99 & 1.74 & - \\
Cremona      & 1:28.22 & 3.73 & 1.31 & 1:29:14\\
Phillip Island & 1:27.49 & 3.10 & 4.35 & - \\
\bottomrule
\end{tabular}
\end{table}

\noindent   
\textbf{Track generalization.} Table~\ref{tab:generalization} shows SPDL-trained agents on three different tracks at iteration 1150. In all cases, the agent completes laps with zero falls and zero wall collisions, demonstrating that the proposed architecture and curriculum scheme generalize across different track geometries. The higher off-track time on Phillip Island is due to brief excursions when the agent navigates a specific chicane at high speed.

\noindent
\textbf{Motorbike generalization.} We transferred the Ducati Panigale V4~R-trained model after 1900 iterations to all four remaining motorbike models without retraining. Different bikes have different specs in terms of weight, power and manoeuvrability, mimicking those of real counterparts and are significantly differnt to drive. Table~\ref{tab:bikes} shows the best lap times. The Kawasaki Ninja ZX-10RR transfers best, trailing the Ducati by only 0.87~s in Barcelona. Honda and Yamaha models show competitive performance, and even outperform the Ducati on Phillip Island due to better handling of a track-specific corner. Only the more aggressive and hard-to-drive BMW M1000~RR struggles on Phillip Island, failing to complete the lap. These results confirm that the learned policy captures generalizable motorbike control priors, not just bike-specific dynamics.

\begin{table}[t]
\centering
\caption{Cross-bike transfer: best single lap times (mm:ss.sss).}
\label{tab:bikes}
\begin{tabular}{@{}lccc@{}}
\toprule
Motorbike & Barcelona & Cremona & Phil.~Island \\
\midrule
Ducati Panigale V4~R (trained) & 1:30.24 & 1:27.80 & 1:26.01 \\
Kawasaki Ninja ZX-10RR         & 1:31.10 & 1:27.98 & 1:28.75 \\
Honda CBR1000RR                & 1:32.34 & 1:28.81 & 1:24.62 \\
Yamaha R1                      & 1:32.53 & 1:29.52 & 1:24.29 \\
BMW M1000~RR                   & 1:33.13 & 1:30.32 & — \\
\bottomrule
\end{tabular}
\end{table}
\subsection{Qualitative results}

\begin{figure}[!t]
\begin{center}
    \includegraphics[width=.6\linewidth]{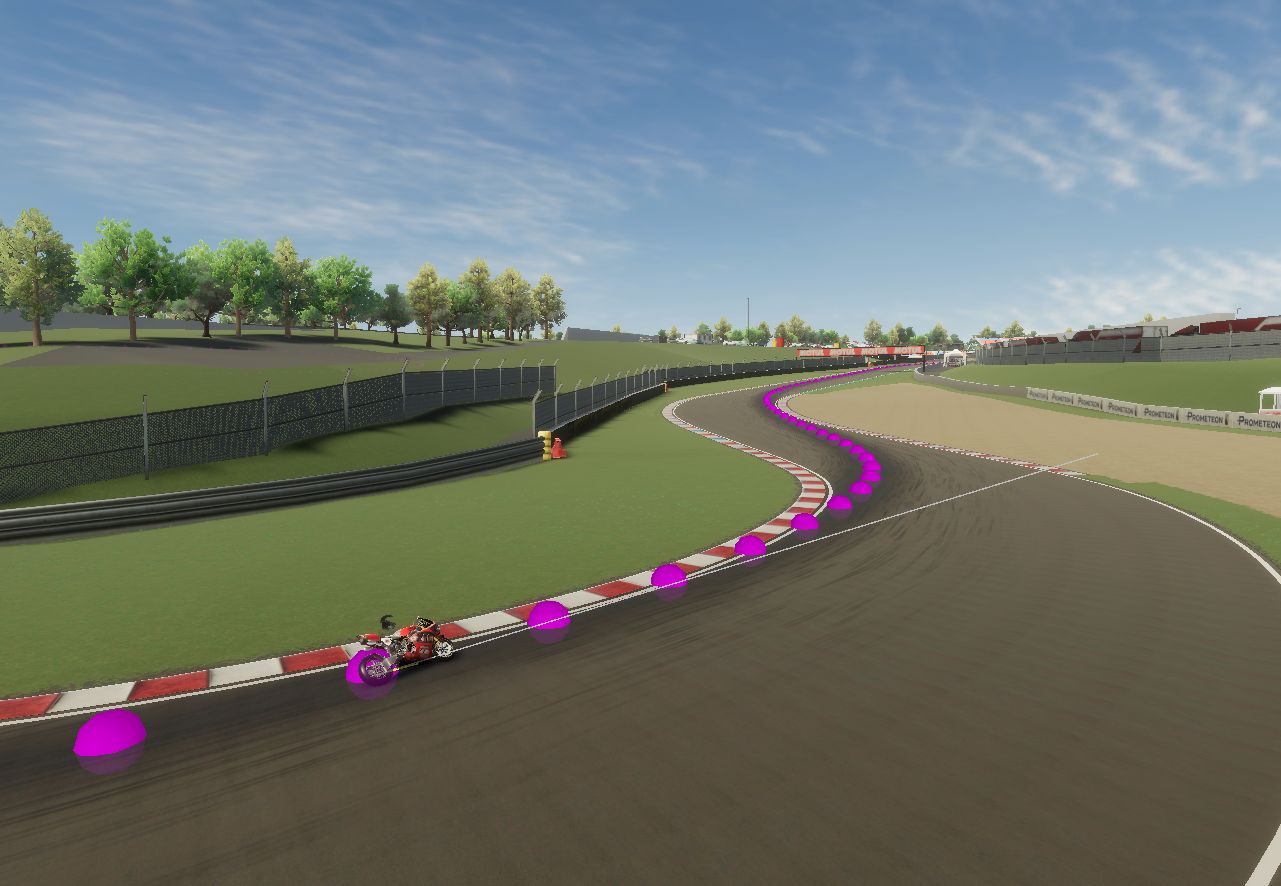}
    \includegraphics[width=.6\linewidth]{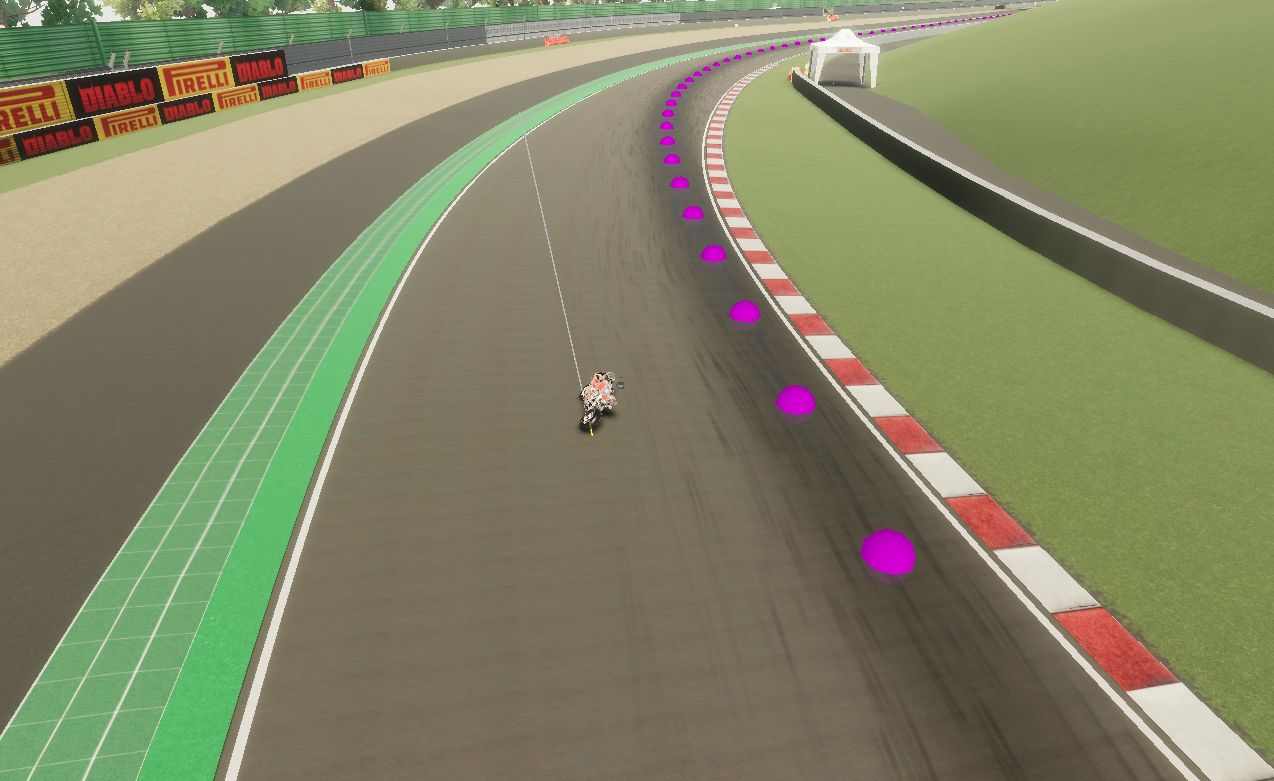}
    \caption{
   Two trajectories followed by the motorbike. In the top one, the bike is able to follow the trajectory in a proper way. In the bottom one, the bike enters too fast into the corner and struggles to follow the ideal trajectory.
    }
    \label{fig:overshoot}
\end{center}
\end{figure}

While early results are promising, a notable limitation of the current system is that the agent prioritizes speed over trajectory precision, overshooting corners. 
An example of this is in Fig. \ref{fig:overshoot}. Against this, future work should incorporate explicit velocity planning signals or trajectory-adherence rewards during high-curvature segments. 

Additionally, the current framework trains a separate agent per track; extending to multi-track generalization — where the agent learns geometry-agnostic driving priors — is an important direction. Finally, the asymmetric actor-critic architecture of \cite{vasco2024super1}, where the critic receives privileged global information during training, could further improve sample efficiency, as could vision-based inputs for a more realistic sensing modality.

\section{Conclusion}

We presented the preliminary results of a Self-Paced Curriculum Reinforcement Learning framework for autonomous superbike racing in VRider SBK. By integrating SAC with SPDL and a motorbike-tailored state-action-reward formulation, our agent learns to race a motorbike in time-trial scenarios, outperforming vanilla SAC in training efficiency and lap time. The framework generalizes across multiple tracks and motorbike models, establishing, to the best of our knowledge, the first RL baseline for autonomous racing bikes. Future works will improve the current framework, testing performance in competitive racing settings. 

\section*{Acknowledgment}

The authors want to thank the European Union’s Horizon 2020 RIA program ELISE GA No 951847.

\bibliographystyle{IEEEtran}
\bibliography{citations}

\end{document}